\pdfoutput=1

\documentclass[11pt]{article}

\usepackage{ACL2023}

\usepackage{times}
\usepackage{latexsym}
\usepackage{graphicx}
\usepackage{booktabs}
\usepackage{colortbl}
\usepackage{subcaption}
\usepackage{multirow}
\usepackage{tabularx}
\usepackage{cuted}
\usepackage{float}
\definecolor{highlight}{RGB}{255,255,204}
\definecolor{darkred}{RGB}{255,150,150}     
\definecolor{lightred}{RGB}{255,200,200}    
\definecolor{darkyellow}{RGB}{255,230,150}  
\definecolor{lightyellow}{RGB}{255,255,200} 
\definecolor{lightgreen}{RGB}{220,255,220}  
\definecolor{darkgreen}{RGB}{180,255,180} 
\usepackage[T1]{fontenc}

\usepackage[utf8]{inputenc}

\usepackage{microtype}

\usepackage{inconsolata}

\title{Knowing What's Missing: Assessing Information Sufficiency in Question Answering}

\author{
  Akriti Jain, Aparna Garimella \\
  Adobe Research, India \\
  \texttt{\{akritij, garimell\}@adobe.com}
}
\begin{document}
\maketitle

\begin{abstract}
Determining whether a provided context contains sufficient information to answer a question is a critical challenge for building reliable question-answering systems. While simple prompting strategies have shown success on factual questions, they frequently fail on inferential ones that require reasoning beyond direct text extraction. We hypothesize that asking a model to first reason about what specific information is missing provides a more reliable, implicit signal for assessing overall sufficiency. To this end, we propose a structured \textit{Identify-then-Verify} framework for robust sufficiency modeling. Our method first generates multiple hypotheses about missing information and establishes a semantic consensus. It then performs a critical verification step, forcing the model to re-examine the source text to confirm whether this information is truly absent. We evaluate our method against established baselines across diverse multi-hop and factual QA datasets. The results demonstrate that by guiding the model to justify its claims about missing information, our framework produces more accurate sufficiency judgments while clearly articulating any information gaps.

\end{abstract}


\section{Introduction}

Large Language Models (LLMs) have demonstrated remarkable capabilities in question answering (QA) \cite{ma2025largelanguagemodelsmeet,lauriola-etal-2025-analyzing,shi-etal-2025-ambiguity,yue2025surveylargelanguagemodel,fischer2024questionlargelanguagemodels}. 
However, a critical limitation undermines their reliability where these models often generate confident-sounding but unfaithful or hallucinated responses when the provided context lacks sufficient information \cite{fadeeva2025faithfulnessawareuncertaintyquantificationfactchecking,ming2025faithevallanguagemodelstay,joren2025sufficientcontextnewlens,ji2023survey}.
\begin{figure*}[t]
    \centering
    \includegraphics[width=0.85\textwidth]{images/hero_image_suff_updated.png}
    \caption{\textbf{LLMs Excel at Factual Verification but Falter on Inferential Reasoning.} 
    (\textbf{Left}) For a fact-based multi-hop question, the model correctly identifies a missing factual link (the location of TCS headquarters) and accurately classifies the context as \textit{Insufficient}.
    (\textbf{Right}) Conversely, for a question requiring simple inference, the model fails. It incorrectly predicts \textit{Insufficient} when asked for the purpose of a business grant, unable to deduce that the scheme's stated function, a scheme which enables local authorities to keep part of business rates-directly serves as its purpose. This highlights a key limitation where models struggle to assess sufficiency beyond literal fact-checking.}
    \label{fig:hero_image}
\end{figure*}

Recent efforts to address this challenge have primarily relied on simple prompt-based strategies that directly ask models to assess whether sufficient information is available \cite{joren2025sufficientcontextnewlens}.  
Such approaches show promise for factual questions, both single-hop (e.g., ``What is the capital of France?'') and multi-hop (e.g., ``Who directed the movie that won Best Picture in 1994?''), where the task reduces to verifying the presence of specific entities or factual chains. However, they frequently fail on inferential questions that require synthesizing information not directly stated in the text (e.g., ``Why did the peace negotiations fail according to the sources?'') as illustrated in Figure \ref{fig:hero_image}. While models correctly identify missing factual links in multi-hop chains, they struggle to assess sufficiency for questions requiring inference. The core issue is that verifying keyword presence differs fundamentally from evaluating whether available facts provide adequate logical foundation for reasoning.

Existing prompting methods typically fall into two categories. 
\textit{Direct sufficiency prompting} asks for an explicit binary judgment (sufficient or insufficient). While effective for fact-checking, this approach proves unreliable when sufficiency depends on whether scattered information collectively supports an inference.
\textit{Answering-based methods} \cite{wang2024llmsknowneedleveraging} infer sufficiency from whether a model provides an answer versus abstaining. However, models often synthesize plausible-sounding answers from inadequate context using parametric knowledge, or hallucinate justifications for abstention that are not grounded in the text.

We hypothesize that a more robust approach for determining context sufficiency is to frame the task around identifying what specific information is missing. 
This reformulation offers two key advantages. First, it forces the model to form a concrete hypothesis about an information gap, which can lead to a more reasoned sufficiency judgment. Second, it produces a granular and actionable output that can be highly beneficial for downstream applications. For example, guiding the next query in an iterative retrieval system to find the necessary context can even improve the QA performance. Recent work on utility-based passage selection has shown that modeling which passages actually contribute to answering a question can significantly improve both retrieval and downstream performance \cite{zhang2025distillingsmallutilitybasedpassage, jain-garimella-2025-modeling, xu2025trainingutilitybasedretrievershared}. Our work complements this line of research by focusing on whether the retrieved context is collectively sufficient, rather than scoring individual passages. 
However, simply asking for missing information may not suffice, as the model may hallucinate non-existent gaps. To address this, we propose an \textbf{Identify-then-Verify} framework that first generates multiple hypotheses about missing information and establishes a consensus, then verifies this consensus against the source text. If the predicted information is indeed missing, the second step would fail to attribute it back, and we use this 
signal as the verification for the first stage's output.

\begin{figure*}[htbp]
\centering
\includegraphics[width=0.95\textwidth]{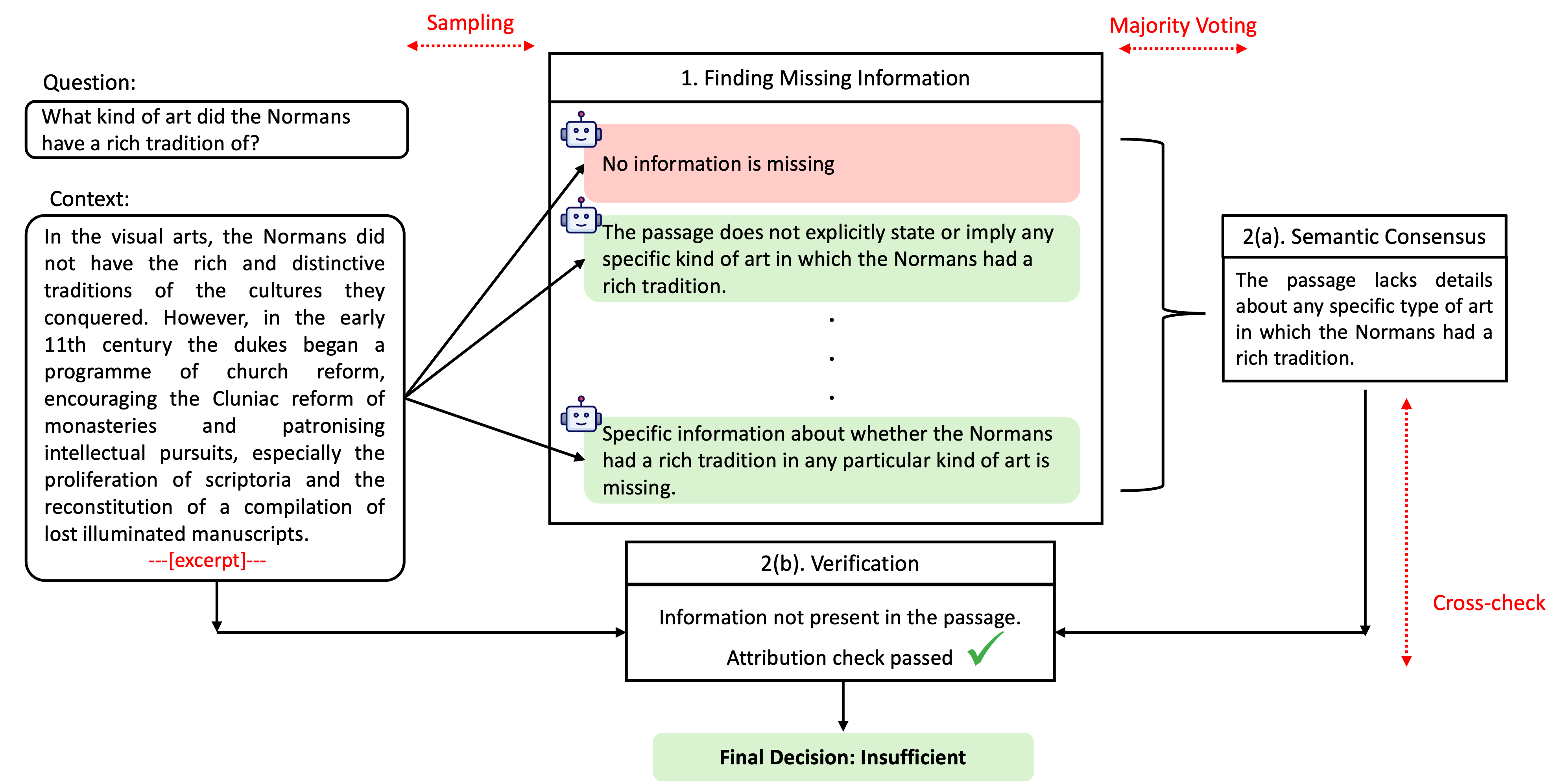}
\caption{\textbf{An Overview of Our Identify-then-Verify Framework.} 
Given a question and context, the framework first queries an LLM multiple times with non-zero temperature to generate a diverse set of hypotheses about missing information (\textbf{Step 1: Identify}). This captures a rich distribution of the model's reasoning. Next, a single Consensus Gap Claim is established from these hypotheses (\textbf{Step 2a}). This claim is then checked against the original context in a final verification step (\textbf{Step 2b}). This verification acts as a critical self-correction mechanism, with the final sufficiency decision based on its outcome.}
\label{fig:method_pipeline}
\end{figure*}

We make three main contributions in this paper.
\noindent {\bf (1)} We establish that information sufficiency prediction remains a challenging problem for open-ended inferential questions by benchmarking existing approaches on various multi-hop and answerability datasets.
{\bf (2)} We propose an Identify-then-Verify framework that reformulates sufficiency assessment as identifying and verifying missing information, leveraging self-consistency and semantic consensus to mitigate hallucination. Our approach can comprehensively identify information gaps, providing actionable outputs for downstream applications.
{\bf (3)} We demonstrate substantial improvements over strong baselines across diverse benchmarks on complex questions both in accuracy and justification alignment. We further establish that sufficiency exists on a spectrum from pragmatic inference to strict literal matching, which our framework accommodates by treating verification strictness as a tunable parameter. 



\section{Related Work and Background}
\subsection{Information Sufficiency and Unanswerable Questions}
For Large Language Models (LLMs) to be truly reliable, they must not only provide accurate answers but also recognize when they lack the information to do so. The study of unanswerable questions has become a widely used framework to investigate this behavior, probing issues such as model hallucination \cite{kim2024detectingllmhallucinationlayerwise,tomani2024uncertaintybasedabstentionllmsimproves}, context faithfulness \cite{ming2025faithevallanguagemodelstay}, and the fundamental question of whether LLMs know when not to answer \cite{saadat2024answerevaluatingpromptsgpt,madhusudhan2024llmsknowanswerinvestigating,deng2024dontjustsayi}.

Successful abstention requires models to reason about both the evidence provided and the uncertainty of their conclusions \cite{kirichenko2025abstentionbenchreasoningllmsfail}. However, a critical distinction must be made: is the model unable to utilize the context correctly, or is the context itself insufficient? \citet{joren2025sufficientcontextnewlens} formalize this distinction by introducing the concept of \textit{sufficient context} and demonstrate that LLMs can directly classify sufficiency for many cases. Our work investigates the limitations of such direct classification approaches, particularly for complex inferential questions, and proposes an alternative formulation based on explicit gap identification.



\subsection{Missing Information Detection}
The concept of prompting a model to identify missing information has been explored for guiding information retrieval as seen in frameworks like MIGRES \cite{wang2024llmsknowneedleveraging}. 
However, the reliability of this method remains a challenge. As demonstrated by benchmarks like AbsenceBench \cite{fu2025absencebenchlanguagemodelscant}, while models excel at recalling surprising information, they still struggle to identify clearly omitted information. On the other hand, recent work on Missing Premise (MiP) problems reveals another failure mode: MiP-Overthinking \cite{fan2025missingpremiseexacerbatesoverthinking}. The study finds that even when reasoning models detect a missing premise early in their thought process, they often fail to commit to this judgment. Instead of abstaining, they fall into excessively redundant and meaningless thinking, generating significantly longer responses while repeatedly attempting to solve an unsolvable problem.
Taken together, these findings show that a single, greedy generation from an LLM is an unstable and insufficient signal. The model might fail to identify the gap entirely, or it might notice the gap but get stuck in a noisy, ineffective reasoning loop.

\subsection{Context Evaluation vs. Context Compression}
A related but distinct line of research focuses on optimizing computational efficiency in context processing. The work of \cite{xie2025knowingstopdynamiccontext} on dynamic context cutoff, for example, seeks to reduce token processing by identifying the earliest point at which an LLM has enough information to answer, effectively performing on-the-fly compression by analyzing internal model states (e.g., attention activations) to discard redundant text.

Our framework differs fundamentally in both goal and approach. Rather than optimizing for efficiency through compression, we perform \textit{evaluation} of context completeness for reliable answering. We analyze the entire, unaltered input to determine whether it contains sufficient information, asking not whether the context can be made smaller, but whether it is adequate in its original form. These approaches are complementary: context compression assumes sufficiency and optimizes processing, while our method explicitly assesses sufficiency to prevent unreliable outputs.

\section{Problem Formulation and Methodology}
We hypothesize that prompting a model for a simple binary prediction (Sufficient/Insufficient) is suboptimal. 
This approach is brittle, particularly for complex inferential questions where the answer does not exist in specific spans but must be inferred or synthesized from multiple portions of the context. 
A model might correctly classify context as insufficient but for an incorrect reason, a nuance lost in binary labels and poorly captured by standard accuracy metrics.
We therefore reformulate the task from a classification problem to a generative one: \textit{What specific information is missing?} This shift forces the model to produce a falsifiable claim, offering a more transparent and interpretable signal of its reasoning.
We propose a structured, two-step Identify-then-Verify framework designed to leverage the generative power of LLMs while mitigating their risks, such as hallucination and inconsistency (Fig. \ref{fig:method_pipeline}).

\noindent \textbf{Step 1: Identify Potential Gaps via Self-Consistency.}

\begin{figure}[t]
    \centering
    \includegraphics[width=0.45\textwidth]{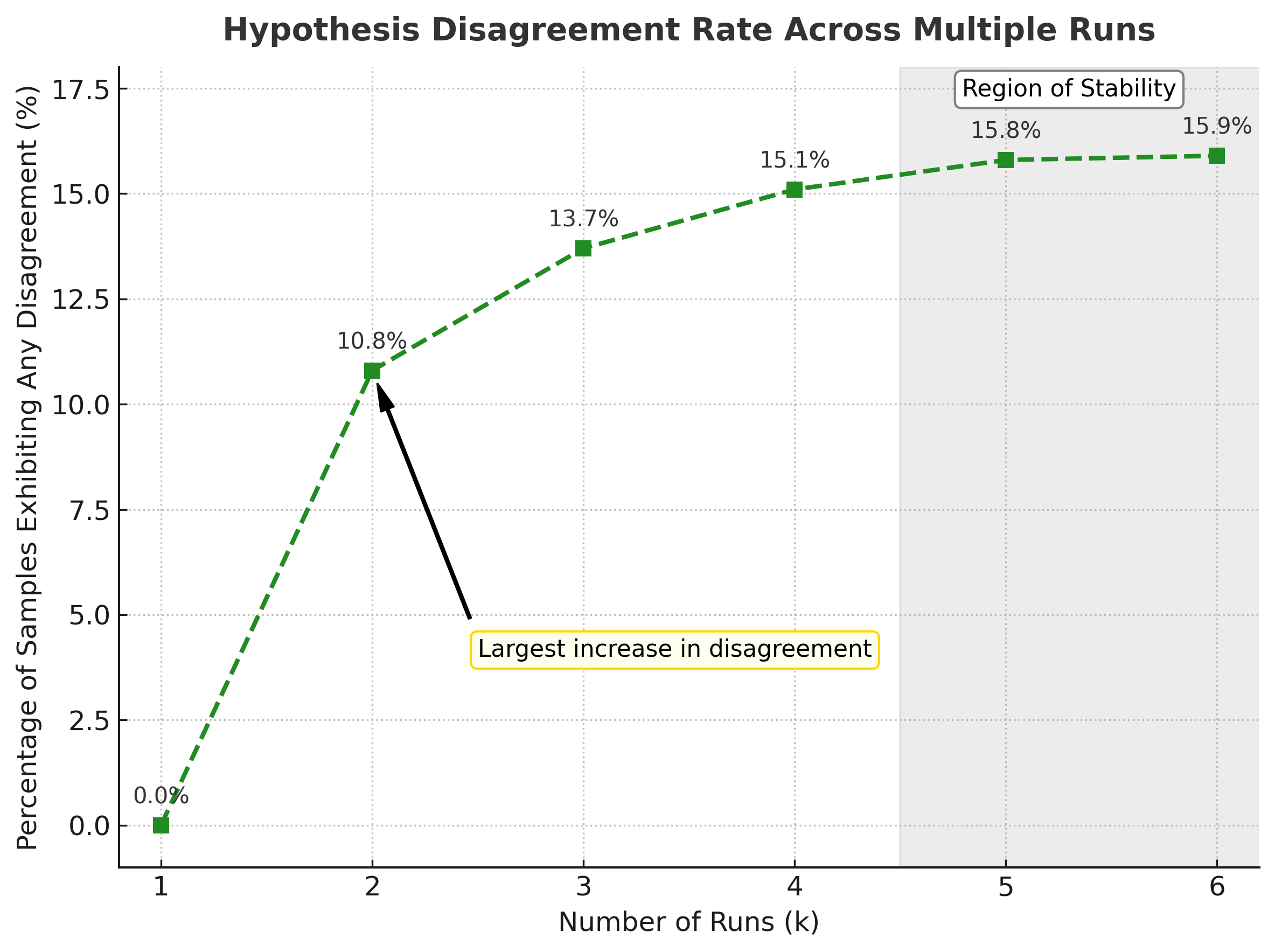}
    \caption{\textbf{Rate of Hypothesis Disagreement Across Multiple Runs.} The plot shows the percentage of samples with conflicting judgments for a given number of runs. The sharpest rise occurs when moving from one to two runs, highlighting the risk of a single pass. The curve flattens significantly after four runs, motivating our choice of $N=5$ as the point of stability.}

    \label{fig:disagreement_plot}
\end{figure}

Rather than asking a model to predict whether context is sufficient, we prompt it to identify \textit{what} information is missing to answer the question. However, LLM responses for missing information detection are brittle, as documented by \citet{fu2025absencebenchlanguagemodelscant}. To quantify this instability, we analyze hypothesis disagreement (Figure \ref{fig:disagreement_plot}): the frequency with which a model produces conflicting judgments about the same question-context pair across multiple runs. Our analysis reveals two key findings. First, disagreement increases sharply from one to two runs, providing strong evidence that single-run generation is unreliable. Table \ref{tab:sufficiency_example} illustrates such disagreement. Second, the disagreement rate stabilizes after 4 runs. Based on these findings, we query the model $N=5$ times with non-zero temperature to generate diverse hypotheses about missing information.

This approach is motivated by recent work demonstrating that aggregating multiple LLM outputs significantly improves reliability and provides robust uncertainty quantification \cite{wang2023selfconsistencyimproveschainthought,yang2025maqaevaluatinguncertaintyquantification,li2025llmsgeneratebetteranswer}. Sampling with non-zero temperature encourages exploration of different plausible token sequences, yielding a distribution of potential information gaps rather than a single deterministic output. This distribution provides a richer signal reflecting the model's assessment of ambiguity. 

\noindent \textbf{Step 2: Establish Consensus and Verify.}
The output of the Identify step is a set of natural language statements, and not classification labels. 
Naive majority voting is therefore inapplicable, as different phrasings may convey the same semantic meaning (e.g., ``the year isn't mentioned,'' ``it lacks a specific date,'' ``the date is missing'').

\noindent {\bf Semantic Consensus.}  We require a more sophisticated aggregation method than a simple majority vote. To facilitate this, we first encode all gap-identifying hypotheses into dense vector embeddings using a sentence Transformer model. From these embeddings, we compute a matrix of cosine similarities. The hypothesis with the highest average similarity to all others is selected as the {\it Consensus Gap Claim}. This ensures we capture the semantic ``center of gravity'' of the model's collective reasoning. If no single idea achieves a minimum semantic similarity threshold, we recognize that the model is highly uncertain and can conservatively default to an insufficient judgment.

\noindent {\bf Attribution Verification.} This Consensus Gap Claim is then subjected to a final, critical attribution check. We initiate a second, independent LLM call with a focused and less ambiguous task: verify if the information specified in the consensus claim is, in fact, absent from the source passage(s). This verification step acts as a self-correction mechanism. By decoupling the exploratory \textit{identification} step from the factual \textit{verification} step, we prevent the model from finalizing a decision based on a hallucinated or misremembered information gap. The final sufficiency determination is based on this attribution check: if the verifier finds the claimed missing information in the context, we classify the context as Sufficient; otherwise, Insufficient.


\begin{table}[t]
\centering
\small
\scalebox{0.9}{%
\begin{tabularx}{\columnwidth}{@{}p{2.2cm}X@{}}
\toprule
\textbf{Question:} & How much did Metallica earn from their 2004 tour? \\
\textbf{Key Context:} & \cellcolor{highlight} Heavy metal band Metallica saw their 2004 earnings rise to \$43.1 million (£23.1 million), thanks to their \textit{Madly in Anger with the World} tour. \\
& [Context excerpt from music industry earnings report\ldots] \\
\textbf{Ground Truth:} & \textbf{Sufficient} \\
\midrule
\textbf{Prediction 1:} & \textbf{Insufficient} \\
\multicolumn{2}{@{}p{\dimexpr\columnwidth-0pt\relax}@{}}{\textit{Reasoning:} The context does not provide sufficient information to directly answer the question. While it mentions Metallica's total earnings in 2004, it does not specify how much came solely from their tour.} \\
\midrule
\textbf{Prediction 2:} & \textbf{Sufficient} \\
\multicolumn{2}{@{}p{\dimexpr\columnwidth-0pt\relax}@{}}{\textit{Reasoning:} The context provides all necessary information to answer the question. It explicitly states that Metallica earned \$43.1m from their 2004 tour.} \\
\bottomrule
\end{tabularx}%
}
\caption{Example of sufficiency disagreement reasoning across runs.}
\label{tab:sufficiency_example}
\end{table}

\section{Experiments}
In this section, we discuss the details of our experimental setup, datasets used, baselines, and evaluation measures.
\begin{table*}[t]
\centering
\small
\scalebox{0.85}{
\begin{tabular}{lrrp{9.5cm}}
\toprule
\textsc{\textbf{Dataset}} & \textsc{\textbf{Before filtering}} & \textsc{\textbf{After filtering}} & \textsc{\textbf{Sample questions}}\\
\midrule
HotpotQA & 7,405 & 2,291 & What is the birthplace of the Senator who represents the first of 62 districts in the State Senate?\\
\midrule
2WikiMultiHopQA & 12,576 & 5,672 & Which film came out first, Kauravar or Beyond The Sunset?\\
\addlinespace[0.3em]
\midrule
MuSiQue & & & \\ 
\quad \textit{Ans} & 2,417 & 1,489 & \multirow{2}{*}{\parbox{9.5cm}{When did the people who captured Malakoff come to the region where Philipsburg is located?}} \\
\quad \textit{Full} & 4,834 & 3,022 & \\
\addlinespace[0.3em]
\midrule
CouldAsk & 4,332 & 4,299 & \\
\quad \textit{BBC} & 278 & 264 & What role did the Lord Chancellor hint might undergo further reforms in the next Labour manifesto?\\
\addlinespace[0.3em]
\quad \textit{Yelp} & 165 & 158 & How is lost and found managed at the Charlotte Convention Center?\\
\addlinespace[0.3em]
\quad \textit{Reddit} & 313 & 301 & Why was the student late to college yesterday?\\
\addlinespace[0.3em]
\quad \textit{Squad v2} & 1,000 & 1,000 & How are the rates of social goods in countries with lower inequality?\\
\addlinespace[0.3em]
\quad \textit{BanditQA} & 2,070 & 2,070 & How long are the Gospel of Matthew writings in the New Testament?\\
\addlinespace[0.3em]
\quad \textit{QA2} & 506 & 506 & Why did the uk ban guns?\\
\addlinespace[0.3em]
\midrule
FaithEval & 2,492 & 2,492 & What is another word for the Earth's upper mantle?\\
\bottomrule
\end{tabular}
}
\caption{Dataset statistics before and after filtering.}
\label{tab:dataset_stats}
\end{table*}

\subsection{Datasets}

We evaluate our framework on two categories of datasets, namely multi-hop question answering and answerability benchmarks.

\noindent\textbf{MultihopQA datasets:} Since these datasets contain questions that are inherently answerable, we create negative (insufficient) samples by removing the designated supporting passages from the context.
        \textbf{HotpotQA} \cite{yang2018hotpotqadatasetdiverseexplainable} features questions that require 2-hop reasoning. Each question is accompanied by 2 gold supporting passages and 8 distractor passages. 
        \textbf{MuSiQue} \cite{trivedi2022musiquemultihopquestionssinglehop} is designed for compositional reasoning. The questions involve 2 to 4 reasoning hops with 20 total passages. We use two settings: MuSiQue-Ans, containing only answerable questions, and MuSiQue-Full, a more stringent and hard version with unanswerable contrast questions designed to prevent disconnected reasoning. 
        \textbf{2WikiMultiHopQA} \cite{ho2020constructingmultihopqadataset} contains 2 or 4-hop questions constructed from Wikipedia, requiring the model to combine information from two different Wikipedia entities or facts to arrive at the answer. Each question is provided with 10 passages.

        
\noindent\textbf{Answerability benchmarks:} These datasets are specifically designed to test models' ability to identify unanswerable questions.
    
         \textbf{CouldAsk} \cite{zhao2024icouldveaskedthat} studies how models handle questions with presupposition errors, where assumptions conflict with or are unsupported by the context. It includes existing datasets (SQuADv2 \cite{rajpurkar-etal-2018-know}, QA2 \cite{kim2023qa2questionansweringquestionable}, BanditQA \cite{gao-etal-2022-simulating}) and three synthetically generated datasets (BBC, Reddit, Yelp) designed to confuse LLMs across diverse domains beyond Wikipedia. The overall performance on these newer datasets (BBC, Reddit, Yelp) for detecting unanswerable questions is often lower than on existing ones, confirming their challenging nature. \textbf{FaithEval} \cite{ming2025faithevallanguagemodelstay} evaluates the contextual faithfulness of LLMs. We focus on its unanswerable setting, where the context is relevant but lacks the specific information needed to answer.

\noindent\textbf{Data Filtering.} Following \citet{dai2025sepermeasureretrievalutility}, we filter out questions that LLMs can answer correctly without any context to mitigate parametric knowledge effects. Such cases would inflate sufficiency detection performance, as models could deem insufficient context as ``sufficient'' by filling gaps from internal memory. Our filtering procedure is as follows: for each question in a given dataset, we first prompt an LLM (GPT-4o) to generate an answer without providing any context. This context-free answer is then evaluated for factual correctness using a second LLM as judge. The evaluator is presented with the original question, the ground-truth answer, and the model's context-free answer, and is tasked with determining if the generated answer is CORRECT, INCORRECT, or UNCLEAR. Any question for which the model's context-free answer is judged as CORRECT is considered ``known'' and is subsequently removed from the evaluation set.
In cases where the ground-truth answers are unavailable, we retain the entire evaluation set. To ensure a fair evaluation, this filtering process is applied to all datasets and for all models, including our baselines.
Applying this filter validates that datasets sourced from static knowledge bases like Wikipedia are particularly susceptible to this issue. 
For instance, we found that on HotpotQA and 2WikiMultiHopQA, the LLM already knew the answers to over $50\%$ of the questions. This rate of knowledge leakage was significantly lower in more recent or specialized answerability benchmarks. 

\subsection{Baseline Methods}
Information sufficiency prediction is a nascent problem without much prior effort; we compare our framework against two recent works that use LLM prompting to assess sufficiency via two distinct strategies:

\noindent \textbf{(1) Sufficient Context Auto-rater}
We use the LLM-based auto-rater from \cite{joren2025sufficientcontextnewlens} by prompting GPT-4o in a one-shot setting to classify whether a given context is sufficient. 

\noindent \textbf{(2) MIGRES} We adapt the framework from \cite{wang2024llmsknowneedleveraging} by reformulating sufficiency detection as a question-answering task. The model is instructed to answer the question based on the provided information. If it cannot, it must respond with ``unanswerable'' and provide a summary of the missing information. We use the unanswerable classification as the sufficiency prediction, representing a strategy where sufficiency is inferred from the model's ability to perform the end task.

\subsection{Implementation Details}
For the hypothesis generation (\textbf{Identify}) step, we use the smaller, open-source Llama-3.1-8B-Instruct \cite{meta2024llama3} model. We generate five hypotheses for each input by setting the sampling temperature to 0.5. To establish semantic consensus, we use all-MiniLM-L6-v2 SentenceTransformer \cite{reimers-2019-sentence-bert} to encode the hypotheses into embeddings. A minimum similarity threshold of 0.3 is used to determine if a clear consensus exists. For attribution verification (\textbf{Verify}), we use GPT-4o \cite{openai2024gpt4technicalreport}, which determines if claimed missing information is present, accepting synonyms, paraphrasing, and reasonable inferences as valid evidence. Notably, our modular pipeline generalizes across model families: the identification stage benefits from sampling diversity (achievable even with smaller models), while the verification task is straightforward enough that both open-source and closed-source LLMs perform comparably (see Section~\ref{sec:model_ablation}). The self-consistency checks run in parallel, and the consensus-finding step is a lightweight similarity computation, making the pipeline practical for real-world RAG applications where skipping sufficiency checks can lead to hallucinations or costly retrieval loops. 

\subsection{Evaluation Metrics}
Since we have gold labels indicating whether the context is sufficient to answer a given question, we report accuracy to validate our approach. Furthermore for the FaithEval benchmark, since we have access to the ground-truth justification of why the context is insufficient, we use this to see how well our method's missing information detection aligns with this justification. We use LLM (GPT-4o) as an auto-evaluator to rate both the explanations on a scale of 1-5, with 1 being not aligned (completely different and unrelated) and 5 being perfectly aligned (both state the same reason).

\begin{table}[ht]
\centering
\small
\setlength{\tabcolsep}{8pt}
\scalebox{0.95}{
\begin{tabular}{lccc}
\toprule
& \multicolumn{3}{c}{\textbf{Accuracy (\%)}} \\
\cmidrule(lr){2-4}
\textbf{Dataset} & \textbf{Autorater} & \textbf{MIGRES} & \textbf{Ours}\\
\midrule
HotpotQA & 83.99& \cellcolor{lightred}84.47& \cellcolor{lightgreen}\textbf{85.24}\\
2WikiMultihopQA & \cellcolor{lightred}87.75& 85.93& \cellcolor{lightgreen}\textbf{89.31}\\
MuSiQue-Ans & 64.85& \cellcolor{lightred}66.96& \cellcolor{lightgreen}\textbf{71.86}\\
MuSiQue-Full & 59.49& \cellcolor{lightred}68.65& \cellcolor{lightgreen}\textbf{72.13}\\
\midrule
\multicolumn{4}{l}{\textit{CouldAsk Benchmark}} \\
BBC & 63.89& \cellcolor{lightred}66.91& \cellcolor{lightgreen}\textbf{78.78}\\
Yelp & \cellcolor{lightred}61.81& 56.36& \cellcolor{lightgreen}\textbf{68.48}\\
Reddit & \cellcolor{lightred}59.49& 58.15& \cellcolor{lightgreen}\textbf{69.65}\\
QA2 & 67.93& \cellcolor{lightred}77.03& \cellcolor{lightgreen}\textbf{78.46}\\
BanditQA & 80.65& \cellcolor{lightred}81.82& \cellcolor{lightgreen}\textbf{83.33}\\
SQuAD v2 & \cellcolor{lightgreen}\textbf{82.87}& \cellcolor{lightred}82.10& 69.50\\
\midrule
FaithEval & \cellcolor{lightgreen}\textbf{78.08}& \cellcolor{lightred}69.04& 57.61\\
\bottomrule
\end{tabular}}
\caption{Comparison of accuracy across QA datasets using different baselines. \textcolor{green}{Green} highlights best accuracies; second best are in \textcolor{red}{red}.}
\label{tab:qa_results}
\end{table}

\section{Results and Discussion}
Our experiments are designed to answer five key research questions. 
\textbf{(RQ1)} How does our framework perform on complex inferential tasks compared to the baselines?
\textbf{(RQ2)} Can the framework identify multiple and diverse types of missing information?
\textbf{(RQ3)} How does the framework adapt to different sufficiency criteria, from pragmatic inference to strict literal matching? 
\textbf{(RQ4)} What is the role of the verification step?
\textbf{(RQ5)} What is the impact of model choice and size on each pipeline stage?

\subsection{RQ1: Performance Across Reasoning Complexity and Inferential Questions}
Table \ref{tab:qa_results} presents our main results. The advantage of our Identify-then-Verify framework grows in proportion to task reasoning complexity. On datasets like HotpotQA and 2WikiMultiHopQA, which are largely extractive in nature, our framework achieves competitive performance with marginal gains over baselines. This aligns with our hypothesis: when sufficiency reduces to verifying specific facts, simple prompting strategies suffice. However, on more complex datasets, the benefits are substantial. For MuSiQue, which features longer reasoning chains, our method yields a notable accuracy gain of up to $5-7$ points. On the CouldAsk benchmark, our method shows superior performance, particularly on BBC, Yelp, and Reddit datasets (10+ point gains). These feature open-ended, inferential questions where the notion of \textit{missing} information is ambiguous. The self-consistency step allows exploration of multiple interpretations, while the verifier consolidates these hypotheses into robust sufficiency judgement.

However, on SQuAD v2 and FaithEval, a simple LLM auto-rater for sufficiency works much better, while our method performs substantially poorly. Upon closer analysis, we note that this points to a much bigger issue in sufficiency prediction, the definition of sufficiency and its subjective nature. Our framework's default setting, which allows for reasonable inferences, is thus penalized by benchmarks requiring strict, literal fact-checking. We address this issue in detail in Section \ref{sec:rq3}.

We also assess statistical significance to validate our gains using McNemar's test, which is designed for comparing classifiers on the same dataset by analyzing their disagreements. Our improvements are statistically significant with $p < 0.008$ on HotpotQA and 2WikiMultihopQA, and $p < 0.0002$ on more complex datasets like MuSiQue and BBC. Per-class precision, recall, and F1 scores are provided in Appendix~\ref{sec:detailed_metrics}.

\subsection{RQ2: Identifying Multiple and Diverse Information Gaps}
A key advantage of our generative formulation is the ability to identify not just whether information is missing, but \textit{what} is missing. Unlike binary classification, our approach produces explicit, comprehensive descriptions of information gaps. Critically, a single hypothesis can identify multiple distinct missing pieces together (Table \ref{tab:multi_gap_example}), and when multiple runs consistently generate similar multi-gap hypotheses, the semantic consensus captures this comprehensive assessment.

\begin{table}[h]
\centering
\small
\scalebox{0.95}{
\begin{tabularx}{\columnwidth}{@{}p{2.2cm}X@{}}
\toprule
\textbf{Question} & Will Hilton scale NoMad internationally as quickly as it scaled Canopy and Tempo? \\
\midrule
\textbf{Context} & Hilton acquired majority interest in Sydell Group to expand NoMad brand globally... [acquisition details, NoMad London flagship mentioned] \\
\midrule
\textbf{MIGRES} & \textbf{Fails to flag insufficiency.} Proceeds to answer ``Yes'' confidently, citing ``Hilton's network of 6,100 hotels'' and naming executives not present in the context. \textit{The model generates a plausible-sounding response using parametric knowledge despite lacking critical information for proper inference.} \\
\midrule
\textbf{Our Method} & \textbf{Correctly identifies insufficiency.} Consensus hypothesis identifies multiple missing pieces needed for inference: (1) specific strategies for scaling NoMad, (2) information about potential obstacles, (3) comparison to Canopy/Tempo expansion timelines. High semantic similarity across runs confirms this comprehensive assessment. \\
\bottomrule
\end{tabularx}
}
\caption{Answering-based methods fail to detect insufficiency and generate confident responses using parametric knowledge, while our method correctly identifies multiple missing information gaps.}
\label{tab:multi_gap_example}
\end{table}


\subsection{RQ3: Adapting to the Spectrum of Sufficiency}

\label{sec:rq3}
We observe that on SQuAD v2 and FaithEval, our default pragmatic configuration is suboptimal (Table \ref{tab:qa_results}). Our error analysis reveals that these benchmarks test strict contextual faithfulness, penalizing any form of inference (Table \ref{tab:faith_eval_dual}).  
For instance, when a question asks for the timing of a CEO's resignation and the context states ``The CEO announced he would step down next year,'' our verifier deems it \textit{Sufficient}, while the benchmark requires a precise date.  
This highlights that information sufficiency is not monolithic but exists on a spectrum, from pragmatic inference to literal fact-checking. Our framework accommodates this spectrum by treating verification strictness as a tunable parameter. For applications requiring faithfulness to exact wording, we simply reconfigure the verification prompt to reject inferences and require explicit matches. The \textit{Identify} step remains unchanged; only the verification criteria shift.

\begin{table}[ht]
\centering
\scriptsize
\captionsetup{justification=justified}
\begin{subtable}{\columnwidth}
\centering
\scalebox{0.95}{
\begin{tabularx}{\columnwidth}{@{}p{2.2cm}X@{}}
\toprule
\textbf{Question:} & What is something that teachers are at a high risk for? \\
\textbf{Key Context:} & Teachers face several occupational hazards in their line of work, including \textcolor{red}{\textit{occupational stress}}, which can negatively impact teachers' mental and physical health, productivity, and students' performance. Stress can be caused by organizational change, relationships with students, fellow teachers, and administrative personnel, working environment, expectations to substitute, long hours with a heavy workload, and inspections. \\
\textbf{Ground Truth:} & \textbf{Insufficient} \\
\textit{GT Justification:} & The sentence mentioning that teachers are at high risk for occupational burnout was removed. Therefore, the modified context no longer supports the original answer. \\
\textbf{Our prediction:} & \textbf{Sufficient} \\
\textit{Our Reasoning:} & The context explicitly states that ''Teachers face several occupational hazards...'', and directly names occupational stress as one of them. \\
\bottomrule
\end{tabularx}
}
\end{subtable}

\vspace{0.6em}

\begin{subtable}{\columnwidth}
\centering
\scalebox{0.95}{
\begin{tabularx}{\columnwidth}{@{}p{2.2cm}X@{}}
\toprule
\textbf{Question:} & Who owns the rail lines in Victoria? \\
\textbf{Key Context:} & Rail transport in Victoria is provided by \textcolor{red}{\textit{several private and public railway operators}}. Major operators include: Metro Trains Melbourne, which runs an extensive, electrified, passenger system throughout Melbourne and suburbs; V/Line, which operates a concentrated service to major regional centres and long-distance services; Pacific National and CFCL Australia, which operate freight services. \\
\textbf{Ground Truth:} & \textbf{Insufficient} \\
\textit{GT Justification:} & The sentence mentioning that the rail lines are 'government-owned' has been removed. Without this information, the ownership of the rail lines in Victoria becomes unclear from the context. \\
\textbf{Our prediction:} & \textbf{Sufficient} \\
\textit{Our Reasoning:} & The passage mentions that rail transport in Victoria is provided by several private and public railway operators, which implies ownership or operational control by these entities. While it does not explicitly state ownership, it is reasonable to infer that these operators are responsible for the rail lines they operate. \\
\bottomrule
\end{tabularx}
}
\end{subtable}

\caption{Two examples from FaithEval, where context is deemed \textbf{Insufficient} because it lacks the precise, literal answer, even though a reasonable inference is possible.}
\label{tab:faith_eval_dual}
\end{table}

With this reconfiguration, our framework's accuracy improves by over $15\%$ on FaithEval (Table \ref{tab:leniency_ablation}). Furthermore, the semantic alignment between our explanations and ground-truth justifications is $4.02$ out of 5, compared to $3.89$ for both baselines, confirming the model identifies gaps for the right reasons. This shows that a single unified method can adapt to different task requirements, from pragmatic QA systems to strict faithfulness benchmarks, with some tuning.

\begin{table}[ht]
\centering
\small
\setlength{\tabcolsep}{8pt}
\scalebox{0.95}{
\begin{tabular}{lccc}
\toprule
& \multicolumn{3}{c}{\textbf{Accuracy (\%)}} \\
\cmidrule(lr){2-4}
\textbf{Dataset} & \textbf{Autorater} & \textbf{MIGRES} & \textbf{Ours} \\
\midrule
SQuAD v2 & 82.87& 82.10& \textbf{84.06}\\
FaithEval & 78.08& 69.04 & \textbf{90.24}\\
\bottomrule
\end{tabular}
}
\caption{Insufficiency accuracy with leniency ablation.}
\label{tab:leniency_ablation}
\end{table}

\begin{table*}[t]
\centering
\small
\begin{subtable}[t]{0.48\textwidth}
\centering
\scalebox{0.8}{
\begin{tabular}{lcccc}
\toprule
& \multicolumn{2}{c}{\textbf{Baselines}} & \multicolumn{2}{c}{\textbf{Ours (+ GPT-4o Verify)}} \\
\cmidrule(lr){2-3} \cmidrule(lr){4-5}
\textbf{Dataset} & \textbf{Auto.} & \textbf{MIG.} & \textbf{Llama 8B} & \textbf{GPT-4o} \\
\midrule
HotpotQA & 83.99 & 84.47 & 85.24 & \textbf{87.60} \\
2WikiMHQA & 87.75 & 85.93 & 89.31 & \textbf{92.60} \\
MuSiQue & 64.85 & 66.96 & 71.86 & \textbf{73.00} \\
BBC & 63.89 & 66.91 & \textbf{78.78} & 77.34 \\
Yelp & 61.81 & 56.36 & 68.48 & \textbf{71.52} \\
Reddit & 59.49 & 58.15 & 69.65 & \textbf{70.60} \\
BanditQA & 80.65 & 81.82 & \textbf{83.33} & 82.95 \\
FaithEval & 78.08 & 69.04 & \textbf{90.24} & 89.62 \\
\bottomrule
\end{tabular}}
\caption{Effect of varying the Identify model.}
\label{tab:identify_ablation}
\end{subtable}
\hfill
\begin{subtable}[t]{0.48\textwidth}
\centering
\scalebox{0.8}{
\begin{tabular}{lccccc}
\toprule
& \multicolumn{2}{c}{\textbf{Baselines}} & \multicolumn{3}{c}{\textbf{Ours (Llama 8B Identify +)}} \\
\cmidrule(lr){2-3} \cmidrule(lr){4-6}
\textbf{Dataset} & \textbf{Auto.} & \textbf{MIG.} & \textbf{GPT-4o} & \textbf{Claude 4} & \textbf{Llama 8B} \\
\midrule
HotpotQA & 83.99 & 84.47 & \textbf{85.24} & 84.52 & 84.64 \\
2WikiMHQA & 87.75 & 85.93 & \textbf{89.31} & 87.87 & 88.00 \\
MuSiQue & 64.85 & 66.96 & \textbf{71.86} & 69.91 & 59.20 \\
BBC & 63.89 & 66.91 & 78.78 & \textbf{79.50} & 71.14 \\
Yelp & 61.81 & 56.36 & \textbf{68.48} & 64.85 & 63.52 \\
Reddit & 59.49 & 58.15 & \textbf{69.65} & 65.00 & 63.67 \\
BanditQA & 80.65 & 81.82 & \textbf{83.33} & 78.89 & 79.03 \\
FaithEval & 78.08 & 69.04 & \textbf{90.24} & 88.20 & 86.00 \\
\bottomrule
\end{tabular}}
\caption{Effect of varying the Verification model.}
\label{tab:verify_ablation}
\end{subtable}
\caption{Model ablation study across pipeline stages. (a) Even a smaller 8B model for identification nearly matches GPT-4o while significantly outperforming baselines. (b) The verification task is straightforward enough that both open-source and closed-source LLMs perform comparably.}
\label{tab:model_ablation}
\end{table*}
\subsection{RQ4: The Role of Verification}
\begin{figure}[h]
    \centering
    \includegraphics[width=0.45\textwidth]{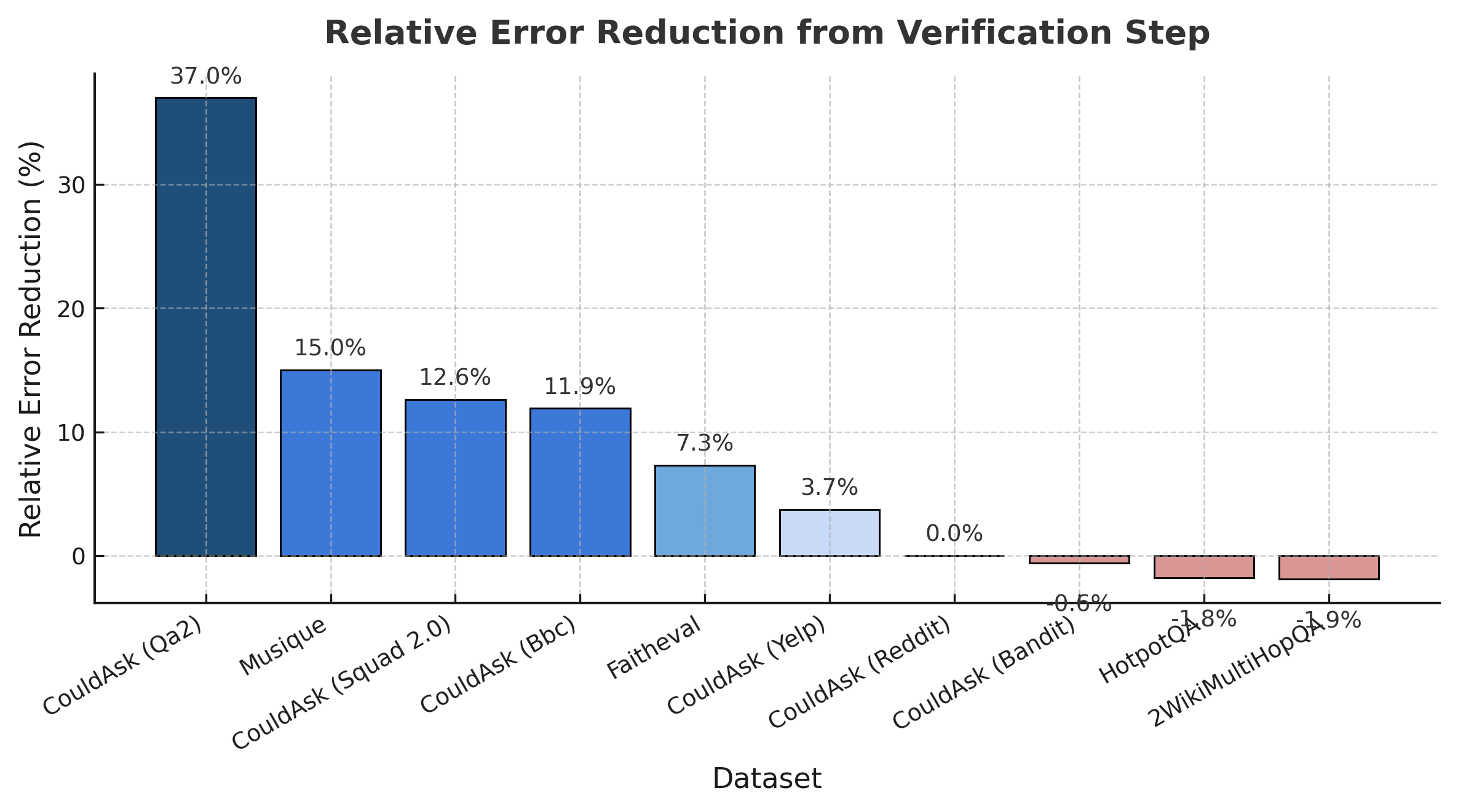}
    \caption{Relative Error Reduction (RER) from the verification step across various datasets. The bars show the percentage of errors from the initial consensus that were corrected.}

    \label{fig:rer_plot}
\end{figure}

To measure the contribution of the verification step, we compare our full Identify-then-Verify framework against an Identify-Only baseline that makes predictions based solely on semantic consensus. Figure \ref{fig:rer_plot} shows the Relative Error Reduction (RER) achieved by adding verification. We use RER rather than absolute accuracy gain as it measures the percentage of errors from the Identify-Only step that were successfully corrected. The verification step yields significant RER across most datasets, with gains as high as $37\%$. This confirms that the verification step is a critical self-correction mechanism, effectively catching hallucinations or misinterpretations from the initial identification stage.

\subsection{RQ5: Model Ablation Study}
\label{sec:model_ablation}
To demonstrate the generalizability of our framework across model families and scales, we ablate both pipeline stages on a representative subset of our evaluation datasets. 

\noindent\textbf{Effect of Identification Model.} Table~\ref{tab:identify_ablation} reveals that even the smaller Llama-3.1-8B-Instruct model generates high-quality hypotheses for the identification stage, with performance nearly matching GPT-4o while significantly outperforming baselines. This validates our design choice of using a cost-effective smaller model for the first stage.

\noindent\textbf{Effect of Verification Model.} Table~\ref{tab:verify_ablation} shows that while fixing identification to Llama-3.1-8B-Instruct, various LLMs (GPT-4o, Claude 4 Sonnet, and Llama-3.1-8B-Instruct) perform comparably on the verification task, all consistently exceeding baselines. This confirms that verification is a straightforward enough task that for both open-source and closed-source models.

Overall, these results validate the effectiveness of our task formulation across diverse model types, allowing practitioners to choose models based on cost or latency requirements. 
\section{Conclusion}

We address robust assessment of information sufficiency for question answering with LLMs. While existing approaches rely on direct binary classification or answering-based detection, both prove brittle for complex inferential questions. We propose an Identify-then-Verify framework that reformulates sufficiency as a generative task: explicitly identifying what information is missing. Our method generates multiple hypotheses via self-consistency, establishes semantic consensus, and performs verification to guard against hallucination. This yields three key advantages: (1) improved accuracy on complex inference questions; (2) comprehensive identification of multiple information gaps, providing actionable guidance for retrieval; and (3) adaptability to different sufficiency criteria, from pragmatic inference to strict literal matching, through simple prompt reconfiguration. A central insight is that information sufficiency varies with application requirements and user needs. By treating verification strictness as a tunable parameter, our framework accommodates diverse requirements. This flexibility, combined with explicit gap identification, makes our approach more reliable and interpretable, paving the way for trustworthy QA systems that know when to abstain and why. 

\label{sec:bibtex}

\section*{Limitations}
Our self-consistency step introduces a trade-off between accuracy and computational cost. While we mitigate this by using a smaller model for hypothesis generation, future research could explore more efficient sampling strategies. 
Further, the adaptability of our framework is currently guided by a manually set strictness definition for sufficiency. A promising direction for future work would be to develop methods that automatically determine the optimal verification setting for a given task, making the framework even more autonomous. Finally, while existing benchmarks serve as valuable initial testbeds for sufficiency assessment, they often do not fully capture the complexity of real-world information-seeking scenarios. Many real-world questions, such as the example in Table \ref{tab:multi_gap_example}, are inherently open-ended and inferential, requiring models to navigate incomplete contexts with multiple potential information gaps. Developing benchmarks that better reflect these characteristics would benefit the broader research community.

\section*{Ethics Statement}
There are no ethical concerns to the best of our knowledge.

\bibliography{anthology,custom}
\bibliographystyle{acl_natbib}
\clearpage
\onecolumn
\appendix
\section{Appendix}

\subsection{Detailed Performance Metrics}
\label{sec:detailed_metrics}
Table~\ref{tab:detailed_metrics} presents per-class precision (P), recall (R), and F1 scores for all methods.

\begin{table}[H]
\centering
\scalebox{0.85}{
\begin{tabular}{llccc|ccc|ccc}
\toprule
& & \multicolumn{3}{c}{\textbf{Ours}} & \multicolumn{3}{c}{\textbf{MIGRES}} & \multicolumn{3}{c}{\textbf{Autorater}} \\
\cmidrule(lr){3-5} \cmidrule(lr){6-8} \cmidrule(lr){9-11}
\textbf{Dataset} & \textbf{Class} & \textbf{P} & \textbf{R} & \textbf{F1} & \textbf{P} & \textbf{R} & \textbf{F1} & \textbf{P} & \textbf{R} & \textbf{F1} \\
\midrule
HotpotQA & Insufficient & 87.7 & 82.0 & 84.7 & 81.8 & 88.7 & 85.1 & 78.9 & 90.5 & 84.3 \\
& Sufficient & 83.1 & 88.5 & 85.7& 87.7 & 80.2 & 83.8 & 88.8 & 75.8 & 81.8 \\
\midrule
MuSiQue-Ans & Insufficient & 67.8 & 82.0 & 74.2& 61.2 & 92.6 & 73.7 & 58.9 & 97.8 & 73.5 \\
& Sufficient & 77.2 & 61.1 & 68.2& 84.8 & 41.3 & 55.6 & 93.7 & 31.9 & 47.6 \\
\midrule
2WikiMHQA & Insufficient & 83.7 & 98.1 & 90.3& 80.4 & 95.1 & 87.1 & 89.1 & 86.1 & 87.5 \\
& Sufficient & 97.7 & 80.9 & 88.5& 94.0 & 76.8 & 84.5 & 86.5 & 89.4 & 87.9 \\
\midrule
BBC & Insufficient & 50.0 & 47.5 & 48.7& 33.3 & 55.9 & 41.8 & 33.3 & 69.5 & 45.1 \\
& Sufficient & 86.0 & 87.2 & 86.6& 85.5 & 69.9 & 76.9 & 88.3 & 62.4 & 73.1 \\
\midrule
Yelp & Insufficient & 49.1 & 51.0 & 50.0 & 39.2 & 74.5 & 51.4 & 43.8 & 82.4 & 57.1\\
& Sufficient & 77.7 & 76.3 & 77.0& 80.9 & 48.2 & 60.4 & 87.0 & 52.6 & 65.6 \\
\midrule
FaithEval & Insufficient & 100.0 & 90.2 & 94.9& 100.0 & 69.0 & 81.7 & 100.0 & 78.1 & 87.7 \\
\bottomrule
\end{tabular}}
\caption{Per-class precision, recall, and F1 scores. Our method generally achieves higher F1 scores, particularly on the Sufficient class where baselines often over-predict insufficiency.}
\label{tab:detailed_metrics}
\end{table}

\end{document}